\pdfoutput=1

\PassOptionsToPackage{numbers, compress}{natbib}

\documentclass[11pt]{article}

\usepackage[preprint]{neurips_data_2024}

\usepackage{times}
\usepackage{latexsym}
\usepackage{hyperref}
\hypersetup{
    colorlinks=true,
    }

\usepackage[utf8]{inputenc} 
\usepackage[T1]{fontenc}    
\usepackage{booktabs}       
\usepackage{amsfonts}       
\usepackage{nicefrac}       
\usepackage{microtype}      
\usepackage{xcolor}         

\usepackage{todonotes}
\usepackage{pgfplots}
\usepackage{multirow}
\usepackage{float}
    
\usepackage{graphicx}

\usepackage[T1]{fontenc}

\usepackage[utf8]{inputenc}

\usepackage{microtype}

\usepackage{inconsolata}

\makeatletter

\newcommand{\ie}{\emph{i.e.}\@ifnextchar.{\!\@gobble}{}}
\newcommand{\eg}{\emph{e.g.}\@ifnextchar.{\!\@gobble}{}}
\newcommand{\etc}{\emph{etc.}\@ifnextchar.{\!\@gobble}{}}

\newcommand{\datalink}{\href{https://www.kaggle.com/datasets/googleai/fsboard}{this link}}
\newcommand{\datasetname}{FSboard}
\newcommand{\datasetlength}{3 million characters}

\newcommand{\baselinecer}{11.1}
\newcommand{\baselineacc}{52.9}

\makeatother

\title{{\datasetname}: Over {\datasetlength} of ASL fingerspelling collected via smartphones}

\author{Manfred Georg$^1$\thanks{ equal contribution~~$\dagger$ equal advising~~$\ddagger$ work conducted while at Google}~, Garrett Tanzer$^{1*}$, Saad Hassan$^{2\ddagger}$, Maximus Shengelia$^{3\ddagger}$, \\ \bf Esha Uboweja$^1$, Sam Sepah$^1$, Sean Forbes$^{4\dagger}$, Thad Starner$^{1*\dagger}$ \\
  $^1$Google, $^2$Tulane University, $^3$Rochester Institute of Technology, \\ $^4$Deaf Professional Arts Network \\
  \texttt{\{mgeorg, thadstarner\}@google.com}}

\begin{document}
\maketitle
\begin{abstract}
Progress in machine understanding of sign languages has been slow and hampered by limited data.
In this paper, we present {\datasetname}, an American Sign Language fingerspelling dataset situated in a mobile text entry use case, collected from 147 paid and consenting Deaf signers using Pixel 4A selfie cameras in a variety of environments.
Fingerspelling recognition is an incomplete solution that is only one small part of sign language translation, but it could provide some immediate benefit to Deaf/Hard of Hearing signers as more broadly capable technology develops.
At $>${\datasetlength} in length and $>$250 hours in duration, {\datasetname} is the largest fingerspelling recognition dataset to date by a factor of $>$10x. As a simple baseline, we finetune 30 Hz MediaPipe Holistic landmark inputs into ByT5-Small and achieve \baselinecer\% Character Error Rate (CER) on a test set with unique phrases and signers. This quality degrades gracefully when decreasing frame rate and excluding face/body landmarks---plausible optimizations to help models run on device in real time.\footnote{We publicly release {\datasetname} at {\datalink} under a CC BY 4.0 license.}

\end{abstract}

\section{Introduction}

The quality of sign language translation, particularly from American Sign Language (ASL) to English, has been steadily improving~\citep{duarte2021how2sign, tarres2023sign, youtubeasl, rust2024privacyaware}, but it is still far from being usable in practice. A body of work on participatory methods for ML~\citep{bragg2019sign,bragg2021the} suggests dividing such an ambitious goal into intermediate milestones that can provide concrete and immediate benefit to the community (\ie, Deaf/Hard of Hearing signers). In this manner, the work starts addressing the community's needs immediately, and the community can drive the direction of future technology.

We focus on the intermediate goal of recognizing fingerspelling as an alternative to smartphone text entry. While full signing for text entry (the proper analogue of speech recognition) is ideal~\citep{Hassan2023-tap-to-sign}, fingerspelling may still be a valuable stopgap due to improved speed or convenience vs. typing on a keyboard. An analogy can be made to gesture keyboards~\citep{zhai2012word}, where the user swipes through the letters of a word as opposed to touching and releasing each letter's virtual key. Even though the system uses pattern recognition to determine which word the typist intended and sometimes returns an incorrect word, many smartphone users prefer such gesture-based keyboards as they feel they can enter text more quickly with the added benefit of requiring one hand instead of two~\citep{reyal2015performance}. Similarly, there is evidence that Deaf signers may find fingerspelling faster or more convenient than current smartphone text entry keyboards \citep{Hassan2023-tap-to-sign}. 

In this paper we present {\datasetname} (\textbf{F}inger\textbf{s}pelling-\textbf{board}, as in ``keyboard''), an ASL fingerspelling dataset situated in a mobile text entry use case. We collect {\datasetname} by creating a domain-appropriate phrase distribution, recruiting 147 paid and consenting Deaf signers through the Deaf Professional Arts Network (DPAN), and having them record one-handed fingerspelled renditions of the phrases using Pixel 4A selfie cameras in a variety of environments. The videos were usually recorded at 1944x2592 pixels and 30 frames per second, though sometimes the resolution varied due to participants accidentally changing settings. At 3.2 million characters in length and 266 hours in duration, {\datasetname} is the largest fingerspelling recognition dataset to date by a factor of $>$10x.

As a simple baseline, we finetune 30 Hz MediaPipe Holistic landmark inputs into ByT5-Small (300M) and achieve \baselinecer\% Character Error Rate (CER) on {\datasetname}'s test set, which features 15 unique signers and no phrase overlap with the train set. We ablate our baseline across several factors like frame rate and exclusion of face/body landmarks which could be used to optimize MediaPipe Holistic's on-device performance, and find that these compromises (in moderation) cause minimal regressions. Qualitatively, the baseline outputs are promising but should be evaluated end-to-end in realistic settings in future work.

We hope that {\datasetname} will help develop text entry methods that start to give signers a more equitable experience with technology, as well as aid in longer term research towards full sign language understanding.

\section{Background}
\label{sec:background}

\begin{figure}
\begin{minipage}{0.3\textwidth}
\centering
\includegraphics[scale=0.5]{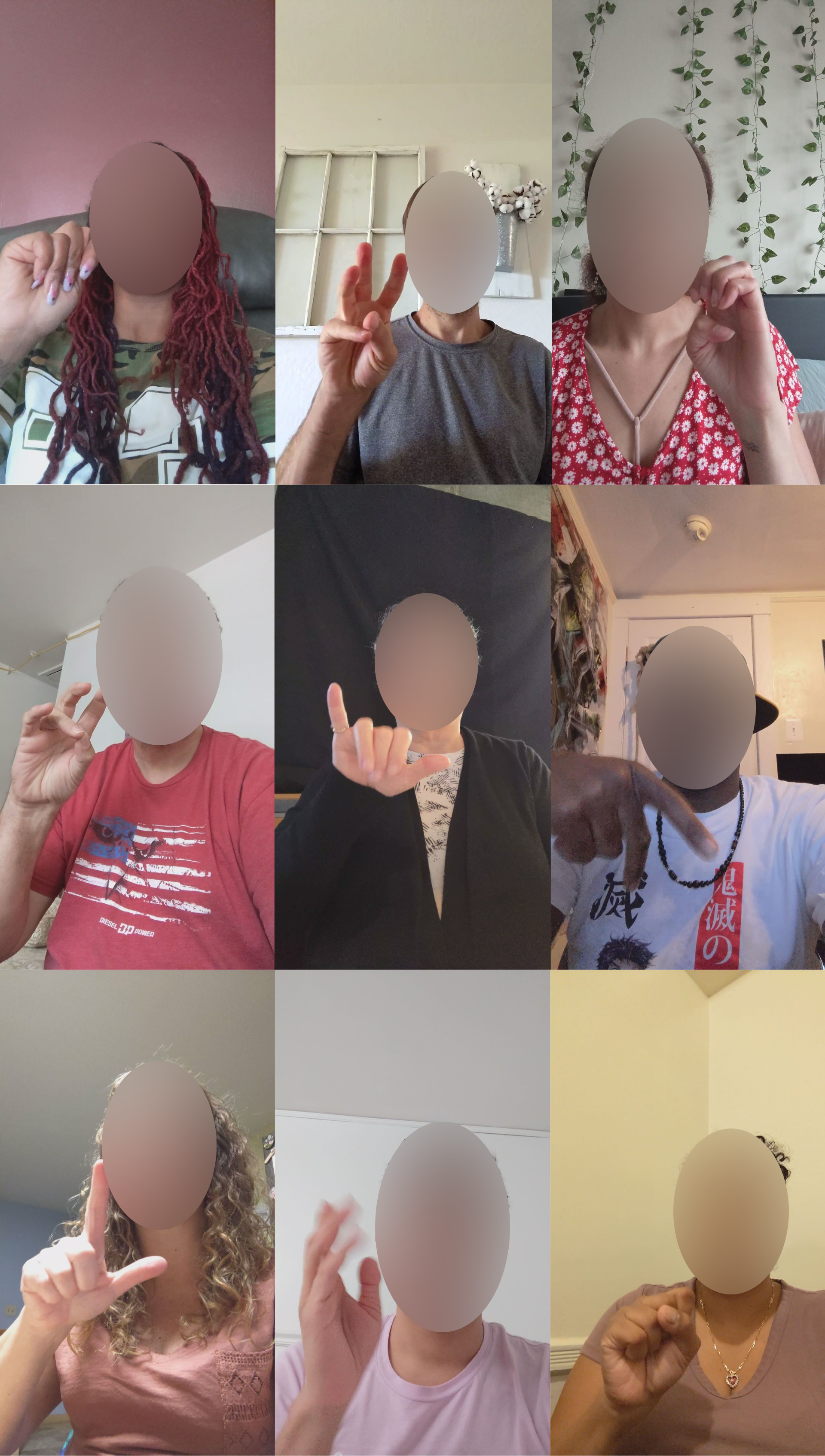} 
\caption{\textbf{A sample of frames from {\datasetname}.} Faces blurred here but not in the dataset.}
\label{fig:image-examples}
\end{minipage}\hfill%
\begin{minipage}{0.66\textwidth}
    \centering
    \vspace{2mm}
    \begin{tabular}{ll}
    \toprule
    \bf MacKenzie~\citep{mackenzie} &  prevailing wind from the east \\
    & elephants are afraid of mice \\\
    & my favorite place to visit \\
    \midrule
    \bf URLs & http://datastudio.google.com \\
    & si.wikipedia.org \\
    & /dfinance/list.asp?id=418/ \\
    \midrule
    \bf Addresses & 9841 gritt hill \\
    & 200ab lake charles \\
    & 24 north 118th place \\
    \midrule
    \bf Phone Numbers &  166-893-6320\\
    & +44-527-848-96-69-05\\
    & +678-92-00-9661 \\
    \midrule
    \bf Names & mohammed kim \\
    & gustavo ho \\
    & clifford davenport \\
    \bottomrule
    \end{tabular}\vspace{1mm}
    \caption{\textbf{A sample of phrases from each category of {\datasetname}.} Addresses, phone numbers, and names are generated randomly; they are not real personally identifiable information (PII).}
    \label{fig:phrase-examples}
\end{minipage}
\end{figure}

According to the United Nations and the World Health Organization, there are over 70 million Deaf and Hard of Hearing people in the     
world~\citep{UN,WHO}. Many use one or more of around 150 sign languages to communicate~\citep{fenlon2015sign}. For example, 
American Sign Language (ASL) is used by about 500,000 people as a primary language in United States alone \cite{one_number}. 

Sign languages are complete, natural languages that can differ from one another significantly even in societies where the same spoken language is used~\citep{abner2024computational,fenlon2015sign}. For example, American Sign Language differs from British Sign Language (BSL) significantly and is instead genetically related to French Sign Language~\cite{abner2024computational}. However, almost all sign languages include a manual alphabet used to represent letters as hand shapes and movements. Often, fingerspelling is used for proper nouns or when introducing new concepts. Fingerspelled terms may also be adapted into the closed vocabulary of a sign language in a process called lexicalization~\cite{johnston2010variation, yanovich2016detection}. For example, the sign for ``job'' in ASL can be seen as a combination of the letters J and B, with the O abbreviated or elided, but the B is in a different orientation from typical fingerspelling. The amount of fingerspelling used in conversational discourse varies depending on the sign language. In ASL, fingerspelling is about 12\%-35\% of signing~\citep{padden2003alphabet,keane2012coarticulation}. 

Some fingerspelling systems are one-handed, such as ASL or Japanese Sign Language (JSL), while others, such as BSL, are two-handed~\cite{power2020evolutionary}. 
Even though meanings may be different, similar hand shapes and movements can be seen across sign languages due to the physical constraints of the hand~\cite{miozzo2022hand}, which suggests that datasets collected for one sign language may offer some transfer to others, or to recognizing handshapes and movements in signing more broadly. This benefit may be especially true for ASL fingerspelling, which belongs to the largest cluster of manual alphabets, the ``French-origin group''~\cite{power2020evolutionary}. For example, many of the letters in the ASL manual alphabet have the same handshape in French, Italian, and German Sign Languages.

There are many works, both informal and academic, which claim to study fingerspelling recognition for ASL and other sign languages but operate on single images. Such efforts are in reality studying handshape classification, with exceptions for fingerspelled letters such as J and Z that incorporate movement.~\citet{ghanem2017survey} provide a survey.
Real-time demonstrations are often slow, with feedback on one letter at a time. These systems ignore the co-articulation effects that occur when recognizing fingerspelling at speed as well as the problem of determining where the space is between two fingerspelled words. In addition, when fingerspelling at speed signers often ``bounce'' or ``slide'' a handshape from inside to outside the body when a letter repeats in a word. We are only aware of the ChicagoFS series of datasets~\citep{fs18iccv,fs18slt,kim2016lexiconfree} that are directly comparable to the effort here for American Sign Language fingerspelling; we provide a detailed comparison to ChicagoFS in Section~\ref{sec:dataset-statistics} below.

\subsection{Fingerspelling for smartphone text entry}

Historically, most sign language recognition systems have had little usefulness or usability for the Deaf community~\citep{bragg2019sign,bragg2021fate, erard2017, hill2020deaf}.
Often the Deaf community is not consulted on the technology being created, nor formative or summative user studies performed. For this work, three of the authors are members of the Deaf community and were integral to the selection of the task, pilot studies and testing, and recruitment of the participants whose primary language is ASL.

Our efforts to create a fingerspelling dataset are motivated by~\citet{Hassan2023-tap-to-sign}, a user study which established the potential benefits of text entry for smartphones based on fingerspelling. It compared an emulated fingerspelling keyboard to normal smartphone typing on Gboard (Android's default keyboard) for 12 Deaf participants and found that fingerspelling was faster than the smartphone keyboard (42.5 wpm vs. 31.9 wpm),
had fewer errors (4.0\% vs. 6.3\%) and had higher throughput (14.2 bits/second vs. 10.9 bits/second). In post-study surveys, 50\% of these Deaf participants preferred fingerspelling for text entry using the emulated recognition system. 

Further adding support that fingerspelling may prove faster than smartphone virtual keyboards, we examined common MacKenzie phrases fingerspelled in the FSboard dataset presented below. Signers averaged around 65 wpm, with some maintaining over 100 wpm. This result is significantly faster than the average smartphone typist at around 36 wpm~\citep{palin2019people} and is consistent with the conversational fingerspelling rates reported in the literature~\citep{quinto2010rates}. 

Texting is often the first use case that comes to mind when thinking about text entry on a smartphone. However, members of the Deaf community have emphasized that fingerspelling to a smartphone may be best suited for entering names or addresses into specific smartphone applications like Google Maps. One can imagine a Deaf signer setting their default keyboard on their smartphone to one that shows both the on-screen keyboard as well as a selfie camera feed that could be used for fingerspelling. In this manner, the signer could easily switch between, or combine, input methods for all applications that require text entry.


\subsection{Community-centered sign language datasets}

PopSign and ASL Citizen are two prior works grounded in focused tasks that could benefit signing communities. Both are  isolated sign recognition tasks (classifying which single sign is present in a given clip). PopSign~\citep{starner2023popsign} is an educational smartphone game intended to help hearing parents of deaf infants practice sign language (and avoid language deprivation \cite{gulati2018language,hall_auditory_2016,hall2019deaf,hall_what_2017, hall_language_2017,humphries2012language,humphries2019support,napoli2015should}).

In the game, the user signs one of five options to select a bubble of a particular color, and these five active options rotate among a library of 250 signs in a way that avoids recognizer confusion. Limiting the number of classes ensures >99\% top-1 accuracy \cite{Bhardwaj2024}. As with FSboard, the PopSign dataset is collected at high resolution (1944x2592) using Pixel 4A smartphone selfie cameras. The PopSign dataset includes over 200,000 clips and 128 hours of video.

ASL Citizen~\citep{desai2023asl}  grounds the sign recognition task in a dictionary retrieval setting, enabling signers to record a video clip of a sign to retrieve its dictionary entry. This task demands a much larger sign vocabulary/number of classes, but retrieval is relatively forgiving because a number of outputs can be returned (\ie, top-N accuracy) from which the user can choose. The authors report a top-1 accuracy of 62\% but a top-10 accuracy of 90\% on a 2731 sign vocabulary. The dataset contains 83,912 videos taken from webcams, and the dataset resolution is often 640x480.

\begin{figure}
\begin{minipage}{0.48\textwidth}
    \centering
    \includegraphics[scale=0.39]{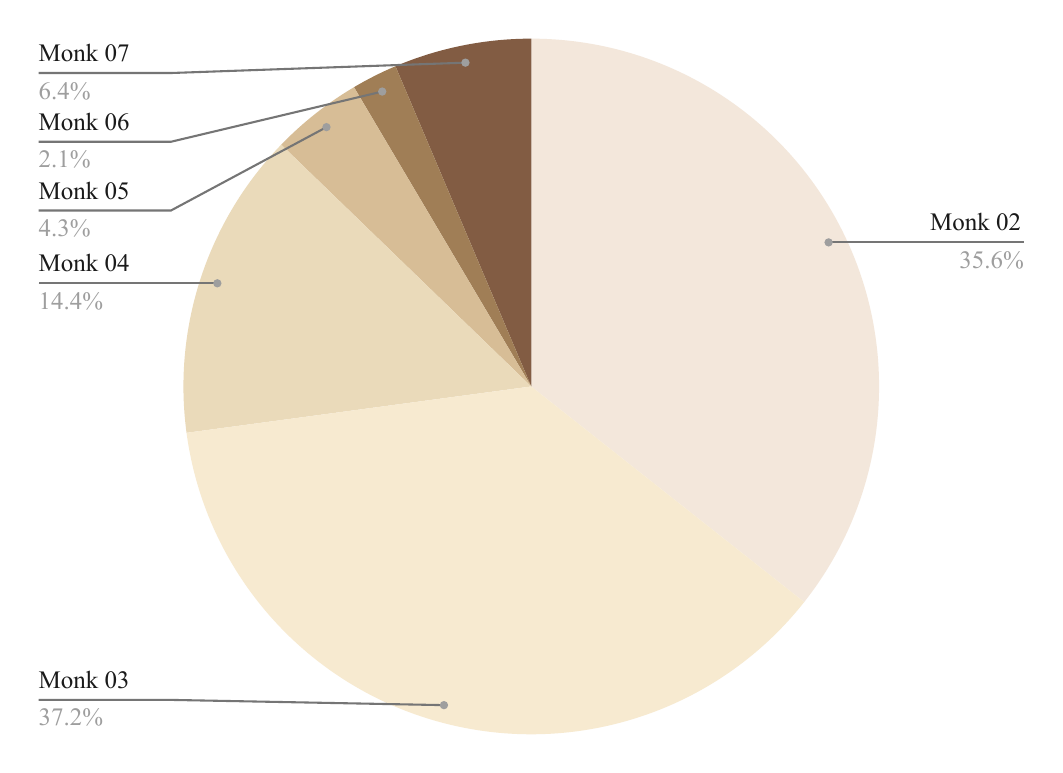} 
    \caption{\textbf{Monk Skin Tone Scale ratings for {\datasetname} participants}, annotated by majority vote of three human raters trained specifically for the skin tone task.}
    \label{fig:skin-tone}
\end{minipage}\hfill%
\begin{minipage}{0.48\textwidth}
    \centering
    \includegraphics[scale=0.39]{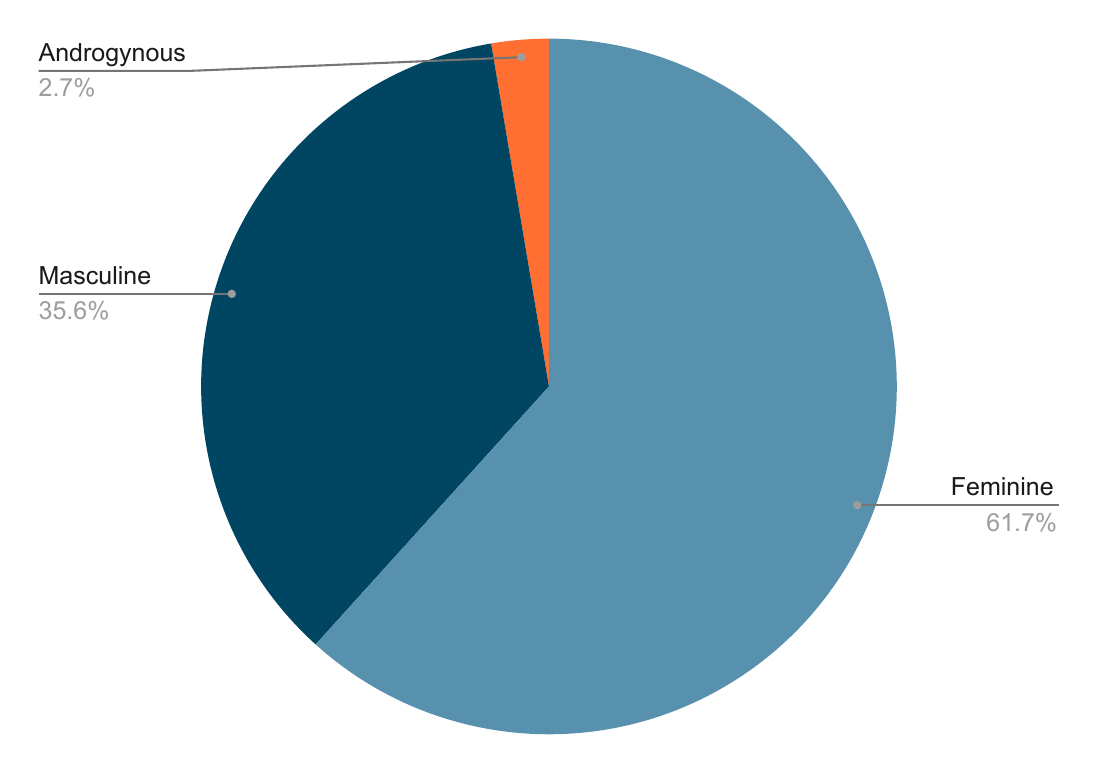}\vspace{-4mm} 
    \caption{\textbf{Perceived gender presentation of {\datasetname} participants}, annotated by human raters. Note that this is not equivalent to gender identity, because it is predicted from the videos rather than self-identified.}
    \label{fig:gender}
\end{minipage}
\end{figure}

\section{The {\datasetname} Dataset}
\label{sec:dataset}

In this section we describe the set of phrases we elicited for {\datasetname} and how the data was collected.

\subsection{Phrases}
\label{sec:phrases}

Numbers should not necessarily be considered fingerspelling, but they are relatively formulaic and often appear in the same contexts as/alongside fingerspelling. There are special signs for various numbers (such as 11, 12, 23, 25, 33, and many more) and specific movements which are generally used while signing numbers, but for practical purposes, the systems for signing cardinal numbers and fingerspelling words are very related, so we choose to include numbers within the scope of our fingerspelling recognition system.

We construct our phrase set with a variety of domains: MacKenzie phrases, URLs, addresses, phone numbers, and names. See Figure~\ref{fig:phrase-examples} for examples from each of these categories.

First, we include the MacKenzie phrase set~\citep{mackenzie}, a classic set of 500 phrases used to evaluate text entry systems. These phrases are intended to be collected multiple times by different signers, to serve as a closed vocabulary testbed for sanity checking methods. The rest of the domains are intended to be unique phrases.

Second, we include randomized URLs using URL parts from a web crawl in April 2022.  The crawled URLs were broken into unique domain name parts and directory parts.  Gibberish parts were removed using a set of manual rules and a simple two character Markov chain trained to recognize URLs a human might want to fingerspell~\citep{gibberish_detector}. Finally, URLs were randomly generated from the parts, sometimes including the protocol identifier (such as \verb!https://!) and sometimes removing the domain name entirely. Some URLs suggest explicit content, which we release as a separate collection of metadata to ensure they are not included in the dataset/models trained on it by default.

Third, we include randomly constructed street addresses. The street names were sampled from the US Census Bureau's 2019 TIGER release \citep{tiger2019}, filtered to remove repetitive, uncommon, and hard to fingerspell names (such as roads that are named by number "Co Rd 87" and "Cr-1601Q4"). Street numbers of 1 to 6 digits were randomly generated with a heavy bias towards 4 digit numbers.  Some addresses had their standard abbreviations expanded ("Lane" for "LN", "Road" for "Rd") while others maintained the abbreviation.

Fourth, we include names randomly generated as combinations of the 1000 most common first and last names in the United States.

Fifth, we include random phone numbers. These include 10 digit US numbers, with and without the "+1", and semi-realistic non-US numbers. The non-US numbers were generated with a valid country code and realistic groupings, but no effort was made to create correct lengths for the country code used.

Although these categories are not exhaustive, they form a solid basis for creating and evaluating a practical fingerspelling system. In the future, we suggest improving the representation of symbols in the dataset and diversity of formats within each category above. More care could also be taken around the inclusion of spaces and instructions to the participants about how to represent spaces in various contexts (either as pauses or, where appropriate, more explicitly with a hand motion). Further clarification of the instructions about whether and how to represent capitalization would also be helpful.

\subsection{Data Collection}

The Deaf Professional Arts Network (DPAN) recruited Deaf signers who use ASL as their primary language to participate in the data collection, of whom 147 completed the task. DPAN shipped loaner Pixel 4A smartphones to them with a custom data collection app installed~\citep{Hassan2023-tap-to-sign}.
This app was a precursor to the open sourced Record These Hands~\citep{recordthesehandsapp} data collection app.

The app showed a phrase as text.  The participant would touch an on-screen button to begin the phrase, then fingerspell the phrase (with the other hand), and finally press another button to advance to the next phrase. There was also a button to be pressed if a mistake was made in fingerspelling the phrase. For most participants, this button did not work properly, leading to issues in data cleanup. A single video was recorded for the entire session with button presses merely recording a timestamp.

The participants were asked to record videos in various settings and circumstances, leading to many different views and backgrounds. Some people wear masks; sometimes the face or portions of the hands are out of frame; sometimes the field of view is at a strange angle recording videos from below. Many participants opted to place the phone down in some way leading to a large variability in distance to participant, motions required for button presses, and timing between button press and the beginning or end of fingerspelling. Particularly for the MacKenzie phrases, most participants ran together the entire phrase with little regard for the separation of words.

\subsection{Data Cleanup}

Due to bugs in the data collection app and some user errors, the timespans recorded above frequently did not capture the actual phrase.
In order to generate reasonable clips we used a bootstrapping method based on our baseline ByT5 model, described in Section~\ref{sec:baselines}.
First the model was trained on a large corpus of YouTube videos with captions.
Next, five models were finetuned to transcribe folds of the noisily bounded fingerspelling data. (Each fold was trained on 4/5 of the data and transcribed the remaining 1/5.)
Since evaluation bias was irrelevant, the dataset splits were performed at the clip level such that every participant was in every split.
Each model was then used to predict text for the remaining 1/5th that it had not yet seen.
Where the model agreed with the clip boundaries and content, the clip was labeled as clean, otherwise the clip was labeled as noisy.
The whole process was repeated two more times (starting each time with a fresh model) using only the clean clips from the previous round.
A significant amount of manual editing and custom rules tailored to each participant were then used to further clean the clips.
There are still some issues with the clip boundaries in the dataset, and it would benefit from further annotation.

\subsection{Dataset Statistics}
\label{sec:dataset-statistics}

We divide {\datasetname} into train, validation, and test splits with unique signers in each split (117, 15, and 15 signers respectively) and no overlap of phrases between splits.\footnote{Due to a bug in the data collection app, sometimes up to 3 signers were prompted to fingerspell the same phrase when we intended to keep the phrases unique. We discarded some data in order to create splits without overlap in signers or phrases.  The signers in each split were chosen so as to minimize the amount of data which needed to be discarded in order to keep each phrase only in one split. This resulted in a 29\% reduction in the number of phrases in the non-MacKenzie portion of {\datasetname}.}

See Figure~\ref{tab:datasets} for statistics about {\datasetname} (and its splits) in comparison to prior fingerspelling recognition datasets. 
At 3.2M characters in phrase length and 266 hours in video duration, {\datasetname} is more than 10x larger than ChicagoFSWild+, the largest prior fingerspelling recognition dataset. The number of sequences in FSboard is only about 3x that of ChicagoFS, reflecting that the average sequence length in {\datasetname} (21.2 characters) is much longer than prior works (5.5 in ChicagoFSWild+) due to the domain. The number of unique signers is also lower (147 vs. 260), since {\datasetname}'s data is created from new participants rather than scraped from the web like ChicagoFSWild(+). {\datasetname} is unique for fingerspelling datasets in that it is recorded from a one-handed smartphone perspective, like PopSign~\citep{starner2023popsign} does for isolated sign classification.

We used trained human annotators to give more visibility into {\datasetname}'s demographic fairness: see Figure~\ref{fig:skin-tone} for Monk Skin Tone Scale~\citep{Monk_2019} ratings and Figure~\ref{fig:gender} for classifications of perceived gender presentation. The dataset has a good amount of variation in skin tone but lacks the lightest and darkest ends of the scale. It is approximately as diverse as OpenASL~\citep{shi2022opendomain} (though the classification system is different), and much more diverse than YouTube-SL-25~\citep{youtubesl25}. Masculine presentation is underrepresented in {\datasetname}, with only 35.6\% of signers. This statistic stands opposite ChicagoFSWild+~\citep{fs18iccv}, which underrepresents feminine presentation to about the same degree; OpenASL and YouTube-SL-25 are essentially at parity.

\begin{figure}
    \centering

    \begin{tabular}{llccccl}
    \toprule
    \bf name & \bf lang & \bf \# seqs & \bf \# chars & \bf \# hrs & \bf \# signers & \bf source \\
    \midrule
    ChicagoFSVid~\citep{kim2016lexiconfree} & ASL & 4K & 21K & <1 & 4 & Lab \\
    ChicagoFSWild~\citep{fs18slt} & ASL & 7K & 38K & 2 & 160 & Web \\
    ChicagoFSWild+~\citep{fs18iccv} & ASL & 55K & 0.3M & 14 & \bf 260 & Web \\
    \midrule
    \bf {\datasetname} (ours) & \multirow{4}{*}{ASL} & \bf 151K & \bf 3.2M & \bf 266 & 147 & \multirow{4}{*}{Smartphone} \\
    \hspace{2mm}\it train && \it 126K &  \it 2.8M & \it 224 & \it 117 \\
    \hspace{2mm}\it validation && \it 12K & \it 0.2M & \it 19 & \it 15 \\
    \hspace{2mm}\it test && \it 13K & \it 0.2M & \it 23 & \it 15 \\
    \bottomrule
    \end{tabular}
    \caption{\textbf{Summary statistics for fingerspelling recognition datasets.}}
    \label{tab:datasets}
\end{figure}

\section{Baselines}
\label{sec:baselines}

We build our baselines on ByT5~\citep{xue2022byt5}, a character-level encoder-decoder language model in the T5 family~\citep{t5}. Following the YouTube-ASL baselines~\citep{youtubeasl}, we linearly project 85 3D MediaPipe Holistic~\citep{lugaresi2019mediapipe, mediapipeholistic} landmarks into the encoder, with one soft token for each frame of input. Unlike YouTube-ASL's baselines, we feed the input frames in at full (30 Hz) frame rate by default, rather than half; we use up to 256 frames of input and 256 characters of output and decode greedily with a beam size of 5. We train each run using 32 TPUv3 with a batch size of 64 and Adafactor optimizer with base learning rate 0.001 for up to 200k steps per run (or convergence), which takes up to 16 hours.

We report character error rate (CER), implemented as length-normalized Levenshtein distance~\citep{levenshtein} using \href{https://www.tensorflow.org/api_docs/python/tf/edit_distance}{TensorFlow's implementation}~\citep{tensorflow2015-whitepaper}, as well as top-1 accuracy (\ie, the fraction of examples that are transcribed perfectly). We select checkpoints based on CER on the validation set. In practice, the sampling-based metrics plateau without apparent overfitting, so checkpoint selection is not especially sensitive.

See Figure~\ref{tab:results} for a full table of quantitative results; see Figure~\ref{tab:qual-examples} for a qualitative sample of outputs. Our baseline achieves \baselinecer\% CER and \baselineacc\% top-1 accuracy on the {\datasetname} test set. This baseline  surpasses the best score (16.4\% CER) set in a Kaggle competition that we hosted based on {\datasetname}~\citep{asl-fingerspelling}, though the participants were limited in terms of model size and runtime. See the \href{https://www.kaggle.com/competitions/asl-fingerspelling/leaderboard}{final leaderboard} for descriptions of many more baseline methods. The main difference is that, as far as we can tell, the top submissions did not use pretrained language models but rather trained models with new, custom architectures on {\datasetname} only.

Our baseline is also substantially better than the 37.7\% CER achieved by ChicagoFSWild+'s baselines~\citep{fs18iccv}.\footnote{We are unable to evaluate our own models on ChicagoFSWild+ due to licensing.} Results obviously cannot be directly compared across test sets in different domains, but it speaks to a combination of our dataset's increased size and choice of domain (including the use of new footage from high-quality selfie cameras, rather than crops of potentially low-resolution web videos). Our baseline results are even better than ChicagoFSWild(+)'s references for human performance at 17.3\% (Wild) and 13.9\% (Wild+), respectively. This result is because, as they note, the datasets are mined as clips within longer signing videos, and the ground truth annotators have access to broader context but the recognizer being tested does not. While semantic context helps to decode fingerspelling in general, the problem is complicated by signers who systematically fingerspell faster and more sloppily when licensed by the discourse context~\citep{patrie}; to some degree attempting to transcribe isolated fingerspelled subclips may be doomed~\citep{reconsideringsentence}. We sidestepped this issue by eliciting new data in the same (isolated) context as the desired task, rather than using clips from preexisting longform data.

We ablate several factors that contribute to our baseline's performance:

\paragraph*{Pretraining.} Finetuning from the pretrained ByT5 checkpoint rather than the randomly initialized architecture makes a massive difference in ultimate performance (\baselinecer\% vs. 33.8\% CER), and also gives much faster convergence (most of the way by 30k steps, vs. at least 200k).

\paragraph*{Model size.} We try scaling from ByT5 Small (300M) to ByT5 Base (580M), but quality decreases. We assume that this dataset is not large enough to warrant the extra modeling capacity, and the model overfits to the target distribution too easily. Models even smaller than ByT5 Small might perform better, but it is the smallest available model in its family.

\paragraph*{Frame rate.} In practice, MediaPipe Holistic is the performance bottleneck for running sign language applications on device, especially for devices that are not quite cutting edge. We ablate frame rate to show that---as expected---quality degrades monotonically with reduced frame rate, but 15 Hz still performs pretty well (\baselinecer\% vs. 11.8\% CER).

\paragraph*{Holistic components.} Likewise, we can improve on-device performance by removing some of the component models of MediaPipe Holistic (which consists of Hands, Pose, and Face). Unlike other aspects of sign language, the meaning of fingerspelling can be read purely from the shape of the hands, which makes this test more principled than it would be in other contexts. Removing the face causes slight degradation (\baselinecer\% to 12.0\% CER), presumably due to the loss of lipreading cues (which different signers use to varying extents), and removing pose has seemingly no effect. It is possible that this gap could grow with more data, if FSboard is not large enough to learn the relevant features.

\begin{figure}
    \centering
    \begin{tabular}{lcc}
    \toprule
    Model & CER ($\downarrow$) & Top-1 Accuracy ($\uparrow$) \\
    \midrule
    Baseline & \bf \baselinecer & \bf  \baselineacc \vspace{2mm} \\

    Pretrained & \bf (\baselinecer) & \bf (\baselineacc) \\
    \hspace{2mm} \it Scratch & 33.8 & 17.9 \vspace{2mm}\\

    ByT5 Small & \bf (\baselinecer) & \bf (\baselineacc) \\
    \hspace{2mm} \it ByT5 Base & 13.3 & 49.1 \vspace{2mm}\\

    30 Hz & \bf (\baselinecer) & \bf (\baselineacc) \\
    \hspace{2mm}\it 30/2 Hz & 11.8 & 51.8 \\
    \hspace{2mm}\it 30/3 Hz & 13.4 & 48.2 \\
    \hspace{2mm}\it 30/4 Hz & 14.6 & 45.1\\
    \hspace{2mm}\it 30/6 Hz & 20.0 & 33.4 \\
    \hspace{2mm}\it 30/8 Hz & 27.1  & 22.2 \\
    \hspace{2mm}\it 30/16 Hz & 64.0 & 0.9 \\
    \hspace{2mm}\it 30/32 Hz & 88.7 & 0.0 \vspace{2mm}\\
    
    Holistic & \bf (\baselinecer) & \bf (\baselineacc) \\
    \hspace{2mm}\it --Face & 12.0 & 50.6 \\
    \hspace{2mm}\it --Face --Pose & 12.5 & 49.7 \\

    \bottomrule
    \end{tabular}
    \caption{\textbf{Character error rate (CER, $\downarrow$) and top-1 accuracy ($\uparrow$) for {\datasetname} fingerspelling recognition baselines.} We provide several ablations with respect to our best-performing baseline (which uses 30 Hz MediaPipe Holistic (Hands+Pose+Face) landmarks finetuned into ByT5 Base): the effect of using pretrained language knowledge vs. training the architecture from scratch on {\datasetname} only, building off ByT5 Small (300M) vs. Base (580M), reducing frame rates, and removing components of Holistic tracking.}
    \label{tab:results}
\end{figure}
\begin{figure}
    \centering
    \begin{tabular}{cll}
    \toprule
    & \bf Target & \bf Prediction \\
    \midrule
    \multirow{13}{*}{\bf Random} & hubert avalos & elbert avalos \\
    & 135-433-9049 & 135-433-9049 \\
    & 870055 sunset creek court & 870055 sunset creek court \\
    & +43-795-19-03-4208 & +43-795-19-03-4208 \\
    & www.rehanfyzio.sk & www.rohanfyzio.sk \\
    & a131003/iandrade82 & a131003/iandrade82 \\
    & eugene or & eugene or \\
    & 331 super chief & 331 super chief \\
    & nashville tennessee & nashvelle tenssee \\
    & /passeig\_de\_maragall & /passigo-de-marsgoll \\
    & https://www.rauschenbach.de & https://www.rauschenbach.de \\
    & 5225 everette mcclerran & 25225 everett mecclar road \\
    & +95-40-860-061-646 & +95-40-860-061-647 \\
    \midrule
    \multirow{6}{*}{\bf Failures} & sparks nv & sparks.net \\
    & www.tttvw.com/lemoyne-pa & www.tt26.com/lenoyne-p \\
    & 2786 lily xing & 78 william g \\
    & 7806 skunk creek road & 378606 skunk creek road \\
    & +54-5828-275-06 & 7158 twp 2303 \\
    & newark new jersey & newton jussey \\
    \bottomrule
    \end{tabular}
    \caption{\textbf{Qualitative examples of our baseline's predictions on the {\datasetname} validation set}. Note that these phrases, like all in the validation set, are unseen in the training set. ``Random'' examples are sampled without cherrypicking, and ``failures'' are a selection of those with the worst errors. For example, for ``tttvw.com'', the fingershapes for ``vw'' and ``two six'' (not twenty six) look identical.}
    \label{tab:qual-examples}
\end{figure}

\section{Limitations}
\label{sec:limitations}

Beyond the intentional limitations in scope of the problem (tackling American Sign Language fingerspelling recognition for a mobile keyboard use case), {\datasetname} has a number of areas for improvement in both data and modeling. 

Our phrase set is a mixture of several relatively narrow domains, and even within those domains the phrases follow some patterns due to the way we synthetically generated them. An independent test set that is based on real queries, rather than being constructed from the same synthetic grammar as the training set (even if the phrases are unique) would give a more robust understanding of the dataset/model's performance and inform future data collection efforts. We have also observed some considerations specific to the mobile keyboard application that were underexplored in our collection because they are not typically important for fingerspelling. One example is capitalization. Signers only distinguish capital from lowercase letters in rare circumstances where it is contextually relevant, but capitalization is more important for text entry and should be elicited intentionally. Special characters and differences in punctuation like hyphens and underscores are also important  but relatively rare/nonstandardized in signing generally.

In terms of modeling, while MediaPipe Holistic handles the vision aspects of fingerspelling recognition in a way that is performant on device, prior work has found limitations in the accuracy of current pose models~\citep{Moryossef_2021_CVPR} (though many of the failure modes relate to interaction between body parts, which is less relevant for fingerspelling). Future work should explore direct modeling of video input, as in~\citet{fs18iccv,fs18slt}, but for the mobile on-device setting.

\section{Conclusion}
\label{sec:conclusion}

In this paper we introduced {\datasetname}, the largest ASL fingerspelling dataset to date by a factor of $>$10x. Informed by a participatory approach that prioritizes real yet tractable needs of Deaf/Hard of Hearing users, {\datasetname} focuses narrowly on a mobile text entry use case to enable signers to fingerspell short phrases and pieces of information as an analogue to faster text entry techniques (as gesture/swipe keyboards are to two thumb typing on smartphones) as opposed to speech recognition (for which full sign recognition would be the analogue). Our baseline achieves \baselinecer\% CER on {\datasetname}'s test set (with unique phrases and signers) and degrades minimally with compromises to frame rate and body tracking that could help maintain real-time on-device performance. We hope that these results (or those achieved by future modeling work on the dataset) will prove high enough quality to be useful in practical applications, and serve as a stepping stone on the way to more generally capable sign language technology.

\section*{Ethics Statement}

The signers who participated in the data collection for {\datasetname} were each paid approximately \$300 for providing 1000 fingerspelled phrases (which typically required 8-12 hours) and consented to their videos being published for a public dataset. Some participants were paid twice, once for the MacKenzie phrase set and once for the addresses, phone numbers, etc. phrase set. 

DPAN is a non-profit Deaf media company whose employees' primary language is ASL. They recruited and consented the contributors to the dataset. While names are not affiliated with any of the videos, it was clear in the consent process that participants’ faces would be identifiable and that the dataset would be public.

While we release the underlying data with faces unblurred because mouthing can be important for fingerspelling recognition, we ask that dataset users blur the signers' faces when including examples in publications (as we do in this paper). Dataset users should not attempt to infer the signers' personal identities or use their likenesses to generate and publish other content (deepfakes).

Given the sociohistorical context surrounding sign language technology and perceptions of fingerspelling, it is important to emphasize that fingerspelling recognition/transcription \textit{is not} sign language translation. Fingerspelling is an important part of ASL, but ultimately \textit{just a part} of the language. Please do not exaggerate the scope of this dataset or task in any follow-up work.

\section*{Acknowledgements and Disclosure of Funding}
We would like to thank all the signers who contributed their data to the project, and Michaela Jitaru and Nathan Qualls at DPAN for their contributions to data collection logistics. Thanks to the Kaggle team (Sohier Dane, Glenn Cameron, Mark Sherwood, Ashley
Chow, and Phil Culliton) for all their advice on how to best create useful datasets. We also thank Chris Dyer for giving feedback on drafts of this paper and Caroline Pantofaru for institutional support. Funding for this dataset is from Google, Inc. DPAN is a non-profit.

\newpage

\bibliographystyle{plainnat}
\bibliography{neurips_data_2023}

\end{document}


\begin{markdown}
# FSboard

FSboard is an American Sign Language fingerspelling dataset consisting of 3.2 million characters and 266 hours of video.  It is 
collected from 147 paid and consented Deaf signers using the selfie camera of the Pixel 4A smartphone in a variety of environments. Using a predecessor for the open source smartphone sign collection app ``Record These Hands,'' participants were prompted to fingerspell short phrases, person names, addresses, URLs, and phone numbers. Video is captured  at 1944x2592 pixels at 30 frames per second (though sometimes the resolution varied due to participants accidentally changing settings).

This dataset is intended for the creation of fingerspelling recognition systems for text entry or educational software for learning fingerspelling,   As conversing in ASL can be 12-35\% fingerspelling, it may also assist in creating a more complete ASL recognition system or ASL generation system. Given the sociohistorical context surrounding sign language technology and perceptions of fingerspelling, it is important to emphasize that fingerspelling recognition/transcription \textit{is not} sign language translation. Fingerspelling is an important part of ASL, but ultimately \textit{just a part} of the language. Please do not exaggerate the scope of this dataset or task in any subsequent work.


#### Dataset Link

The dataset will be hosted on Kaggle \href{https://www.kaggle.com/datasets/garretttanzer/fsboard}{https://www.kaggle.com/datasets/garretttanzer/fsboard} or a similar link updated before the camera-ready deadline. Croissant metadata is available on Kaggle.

We release the dataset under a CC BY 4.0 license and bear all responsibility in case this license is inappropriate. Note that Creative Commons licenses do not supersede other rights such as right of publicity.

#### Data Card Author(s)

- Thad Starner (Owner)
- Garrett Tanzer (Contributor)

## Authorship
### Publishers
#### Publishing Organization(s)

- Google
- Deaf Professional Arts Network

#### Industry Type(s)

- Corporate - Tech
- Not-for-profit - Tech 

### Dataset Owners

#### Contact Detail(s)

- **Dataset Owner(s):** Manfred Georg, Thad Starner
- **Affiliation:** Google
- **Contact:** mgeorg@google.com, thadstarner@google.com

#### Author(s)

- Manfred Georg, Google
- Garrett Tanzer, Google
- Saad Hassan, Tulane University
- Maximus Shengelia, Rochester Institute of Technology
- Esha Uboweja, Google
- Sam Sepah, Google
- Sean Forbes, Deaf Professional Arts Network
- Thad Starner, Google

### Funding Sources
#### Institution(s)

Google

#### Funding or Grant Summary(ies)



Google funded the Deaf Professional Arts Network to collect the dataset with the expectation of the dataset becoming public.

**Additional Notes:** None

## Dataset Overview
#### Data Subject(s)

- Sensitive data about people (videos of people)
- Non-sensitive data about people
- Synthetically generated text data

#### Content Description

This dataset was collected from August 2021 to March 2023.

#### Dataset Snapshot

Size of Dataset: 266 hours 

Number of Signers: 147

Number of Labels: 3.2 million characters

Number of Instances: 151,000 phrases

Average Labels Per Instance: 21.2 characters average per phrase (28.8 MacKenzie phrase; 18.6 addresses/phone/URL/etc.)

Types of Instances: SMS-style phrase (MacKenzie), randomly generated URLs, street addresses, common names (in the United States), phone numbers

Labeled Classes: 74 consisting of 10 digits, space, 11 symbols, 26 letters, 26 capitals (optionally indicated by signers)




%
%
%



### Sensitivity of Data
#### Sensitivity Type(s)

- User Content
- User Metadata
- User Activity Data
- Identifiable Data
- S/PII

#### Field(s) with Sensitive Data

considerations. -->

**Intentional Collected Sensitive Data**

(S/PII were collected as a part of the
dataset creation process.)

Participant Video: Video of participant (upper body captured)       

Participant Sign: Video of participant performing continuous fingerspelling






#### Security and Privacy Handling

**Method:** Participants were given a consent form. They were only allowed to record after providing consent for the following: 

"The app will collect video and photographic images of Your face, torso, hands, and whatever is in view of the camera(s) along with associated camera metadata (such as color correction, focal length, etc.). ... Beyond the video, the following data may be recorded:
 
- The details of each Task, such as the type of Task that was done, signing certain words, or performing specific actions as instructed
- Date and time information associated with the Tasks
- Self identified gender
- Self identified age range
- Self-identified ethnicity
- Self assessed sign language proficiency
- Signing style information (such as general location where You learned, type of sign learned, age range when you started learning, signing community You are most closely associated with, etc)

As described earlier, if you consent, we will use photos or video clips where your face can be identified. We may use identifiable photos or video clips of you in written or oral presentations about this work and in publicly available on-line databases."

#### Risk Type(s)
dataset: -->

- Direct Risk
- Residual Risk

#### Risk(s) and Mitigation(s)

The direct risk involves participants' visual features (their face and body) being linked to their full name. To mitigate this risk, we use anonymized user IDs to identify users. There is still some residual risk. Participants may still be identified using their faces alone. This risk is unavoidable with video data. We have participants sign consent forms acknowledging that they are creating a dataset intended for public use.

### Dataset Version and Maintenance
#### Maintenance Status

**Regularly Updated** - New versions of the dataset
have been or will continue to be
made available.

#### Version Details

**Current Version:** 1.0


**Release Date:** 09/2024

#### Maintenance Plan

**Versioning:** Major updates will be released as a new version, incremented to the nearest tenth from the previous version. For example, if the current version is between 1.0 and 1.09, then a major update will\
 be released as version 1.1. Major updates include the addition of new users and/or new categories of phrases. Minor updates are covered below.         

**Updates:** If there are missing/extraneous/erroneous videos (error cases described below), any fixes will be released as a new version, incremented by 0.01. E.g. if the current version is 1.0, then any minor updates will be released as 1.01.                                                           

**Errors:** Errors in the dataset include incorrectly labeled videos, missing videos, or extraneous videos. Missing videos include videos that the participant recorded but were not included in the final release. Extraneous videos include videos that only have partial input or no sign at all, but were included in the final dataset.                               

**Feedback:** We will accept feedback to the authors' emails.

#### Next Planned Update(s)

**Version affected:** 1.0

**Next data update:** TBD

**Next version:** 1.1

**Next version update:** TBD

#### Expected Change(s)


**Updates to Data:**  The current dataset includes capitalization but some signers did not indicate capitals explicitly. Often, the same symbol would be signed with a variety of hand shapes, which are not detailed in the annotation. Future annotation improvements may address these shortcomings. While the same 500 MacKenzie phrases were collected repeatedly, that was not the intention for the URLs, names, etc. About 1 million more characters were collected but are not released yet as they unintentionally repeated previous prompts. These may be released at a future date.

## Example of Data Points
#### Primary Data Modality

- Video Data

#### Typical Data Point

A typical data point includes the signer fingerspelling a prompt, with little to no empty space (i.e. no motion) at the beginning or the end of the video. The average video is six seconds in length.

#### Atypical Data Point

An atypical data point may be improperly clipped/realigned, so the clip contains contains no fingerspelling, only part of the desired fingerspelled phrase, an erroneous version of the phrase and then a redo, or the wrong fingerspelled phrase.

## Motivations \& Intentions
### Motivations
#### Purpose(s)

- Research
- Production
- Education

#### Domain(s) of Application

`Educational Technology`, `Accessibility`, `Sign Language Recognition`, `Machine Learning`, `Computer Vision`, `Fingerspelling Recognition`, `American Sign Language`

#### Motivating Factor(s)

- Text Entry
- Teaching Sign
- Developing Educational Technology
- Developing Sign Language Recognition

The MobileHCI 2023 paper ``Tap to sign: Towards using american sign language for text entry on smartphones'' demonstrated that Deaf signers fingerspelled significantly faster, on average, than using standard smartphone virtual keyboards when performing text entry tasks. Data collected here suggest that speeds could average as much as twice as fast. In addition, due to dexterity limitations, some older Deaf signers have difficulty with mobile phone text entry but can fingerspell fluently. Our dataset is collected on mobile phones and is designed to facilitate text entry on smartphones for the Deaf. 
Additionally, we foresee the use of this dataset for creating educational technology for helping those who wish to learn sign language gain speed and fluency when fingerspelling. Finally, fingerspelling is an integral part of sign in general, and this dataset may be useful in working towards a conversational sign language recognition or generation system. 

### Intended Use
#### Dataset Use(s)

- Safe for research use

#### Suitable Use Case(s)

**Suitable Use Case:** ASL fingerspelling recognition and generation

**Suitable Use Case:** ASL text entry

**Suitable Use Case:** ASL educational software

**Suitable Use Case:** Pretraining for fingerspelling tasks in other sign languages with a similar alphabet

In general, the data can be used for ASL fingerspelling recognition/generation and related downstream applications.

#### Unsuitable Use Case(s)

**Unsuitable Use Case:** Full Sign Language Recognition/Generation

**Unsuitable Use Case:** Full Sign to English or English to Sign Translation

By itself, this data is not sufficient for continuous sign language recognition/generation or sign to spoken language translation.

#### Research and Problem Space(s)

This dataset is primarily intended to create a fingerspelling recognition system suitable for text entry.

#### Citation Guidelines

Please cite FSboard as follows:

```
@misc{fsboard,
title={FSboard: Over 3 million characters of ASL fingerspelling collected via smartphones},
author={Georg, Manfred and Tanzer, Garrett and Hassan, Saad and Shengelia, Maximus and Uboweja, Esha and Sepah, Sam and Forbes, Sean and Starner, Thad},
year={2024}
}
```

## Access, Rentention, \& Wipeout
### Access
#### Access Type

- External - Open Access

#### Documentation Link(s)

- Dataset Website URL: https://www.kaggle.com/datasets/garretttanzer/fsboard

### Retention
#### Duration

The dataset will be available for at least 5 years.

## Provenance
### Collection
#### Method(s) Used

We collected data from an open-source mobile recording app, now called ``Record These Hands.'' The app presents 10 phrases for recording in a single recording session. The entire session was captured on video, but the sign recordings happened during specific time intervals. The participants were presented a phrase to record. They then tapped a record button to record themselves signing and tapped again to finish signing. The timestamps corresponding to the recording intervals for each sign were saved in a separate file.

**Collection Method:** Record These Hands App

**Platform:** [Platform Name], Google Pixel 4a

**Dates of Collection:** [09 2021 - 03 2023]

**Primary modality of collection data:**

- Video Data

**Update Frequency for collected data:**

- Static

#### Source Description(s)

Participants were recruited by DPAN. Participants are Deaf signers who use ASL as their primary language.

#### Collection Cadence

**Static:** Data was collected once or twice from single or multiple sources.

#### Data Processing

**Collection Method:** Record These Hands

**Description:** We split session recordings from Record These Hands using python. The resulting split videos were named following this convention: "<participant id>-<phrase id>-<recording start time>-<recording index>.mp4". 

**Tools or libraries:** Python, FFmpeg

### Collection Criteria
#### Data Selection

Due to bugs in an early version of the data collection app and some user errors, the timespans recorded above frequently did not capture the actual phrase.
In order to generate reasonable clips a bootstrapping method was used with a ByT5 model.
First the model was trained on a large corpus of YouTube videos with captions.
Next 5 models were finetuned to transcribe fingerspelling data using 4/5ths of the clips with lots of incorrect bounds.
Since evaluation bias was irrelevant, the dataset splits were performed at the clip level such that every participant was in every split.
Each model was then used to predict text for the remaining 1/5th that it had not yet seen.
Where the model agreed with the clip boundaries and content, the clip was labeled as clean, otherwise the clip was labeled as noisy.
The whole process was repeated 2 more times (starting each time with a fresh model) using only the clean clips from the previous round.
A significant amount of manual editing and custom rules tailored to each participant were then used to further clean the clips.
There are still significant issues with the clip boundaries in the dataset and it would benefit from further annotation.

#### Data Inclusion

See above.

#### Data Exclusion

See above.

### Relationship to Source
#### Use \& Utility(ies)

FSboard is intended to train a real-time fingerspelling text entry system for use on smartphones.

#### Benefit and Value(s)

A fingerspelling text entry system may prove significantly faster than currently smartphone text entry for the Deaf; may be preferred by Deaf signers; and may cause less discomfort.  

#### Limitation(s) and Trade-Off(s)

FSboard only focuses on fingerspelling, not full sign language recognition.

## Human and Other Sensitive Attributes
#### Sensitive Human Attribute(s)

- Language
- Culture
- Age
- Gender
- Ethnicity

#### Intentionality

**Intentionally Collected Attributes**

Language: The signers were selected to use ASL as their primary language.

Culture: The signers were selected to be culturally Deaf.

**Unintentionally Collected Attributes**

Human attributes were not explicitly collected as a part of the dataset
creation process, but given that the data includes videos of the participants (including their face), attributes like age, gender, ethnicity, etc. can be predicted using additional methods.

#### Rationale

We intended to collect continuous fingerspelling data; hence videos of the participant's signing were collected. The collected attributes (both intentional and unintentional) may be inferred (though not always accurately) from the videos.

#### Source(s)

We had three professional raters rate the perceived gender and Monk skin tone for FSboard participants. The majority assignment was used.
Please see the main paper for details and distributions.

#### Limitation(s) and Recommendation(s)

Limited scope: FSboard is not intended to address full sign language recognition/translation. Its focus is fingerspelling.

Re-identification: Do not combine FSboard with other datasets in order to re-identify participants. Blur faces when including images in publications or demonstration videos.

#### Risk(s) and Mitigation(s)

The direct risk with this type of video data is with the participant's identity being revealed. For this reason, we use anonymous identifiers. There is still some residual risk with the participant being identified through their faces alone. These participants have signed a consent form (given in the Data Sensitivity section) to address this concern.

### Use in ML or AI Systems
#### Dataset Use(s)

- Training

- Testing

- Validation

- Development or Production Use

- Fine Tuning

## Annotations & Labeling

- Annotation Target in Data

- Machine-Generated

- Human Annotations (Expert)

#### Annotation Characteristic(s)

Besides the automatic affiliation of the phrase with the corresponding video of fingerspelling, we were interested in knowing whether we were collecting a diverse set of skin tones and genders. 

We had three professional raters rate the perceived gender and Monk skin tone for FSboard participants. The majority assignment was used.
Please see the main paper for details and distributions.


## Known Applications \& Benchmarks

Fingerspelling recognition

#### Evaluation Result(s)

See the paper for our baselines on fingerspelling recognition using MediaPipe Holistic and a finetuned ByT5 Small. We achieve 11.1\% CER and 52.1\% top-1 accuracy on the test set, which has nonoverlapping signers and phrases with respect to the training set.

#### Expected Performance and Known Caveats

FSboard has a relatively limited domain, so depending on the method models trained on it may not generalize to other domains.

## Terms of Art

#### ASL
American Sign Language

#### Fingerspelling
A system within various sign languages for spelling out words using a manual alphabet. This is just one small component of sign languages.

## Reflections on Data

Some of the randomly generated URLs collected in the dataset may be offensive to some parties. We do not include phrases that our automatic classifiers deem explicit in the main dataset metadata, but rather release it as a separate file so that it must be used consciously.

\end{markdown}